\documentclass[letterpaper, 10pt, twocolumn]{article}
\usepackage{clean}
\shorttitle{Robotic Skill Diversification via Active Mutation of Reward Functions}

\usepackage[caption=false,font=footnotesize]{subfig}
\usepackage{stfloats}

\usepackage{verbatim}

\usepackage{graphicx}
\usepackage{subfig}
\usepackage{cite}

\usepackage{graphics} % for pdf, bitmapped graphics files
\usepackage{epsfig} % for postscript graphics files
\usepackage{bm,times} % assumes new font selection scheme installed
\usepackage{amsmath} % assumes amsmath package installed
\usepackage{amssymb}  % assumes amsmath package installed
\usepackage{amsfonts}
\usepackage{verbatim} %comment
\graphicspath{{figs/}}
\usepackage{epstopdf}
\usepackage{stfloats}

\usepackage{float}
\usepackage{color}
\usepackage{url}
\usepackage{verbatim}

\usepackage{hyperref}

\usepackage{float}

\usepackage{url}

\begin{document}

\title{\LARGE \bf
Robotic Skill Diversification via Active Mutation of Reward Functions in Reinforcement Learning During a Liquid Pouring Task
}

 \author{Jannick van Buuren$^1$, Roberto Giglio$^{2}$, Loris Roveda$^{2,3}$, and Luka Peternel$^1$
 \thanks{$^1$Cognitive Robotics, Delft University of Technology, Delft, The Netherlands (Corresponding author e-mail: l.peternel@tudelft.nl)}
 \thanks{$^2$Department of Mechanical Engineering, Politecnico di Milano, Italy}
 \thanks{$^3$Istituto Dalle Molle di studi sull'intelligenza artificiale (IDSIA USI-SUPSI), Scuola universitaria professionale della Svizzera italiana, DTI, Lugano, Switzerland}
 }

% make the title area
\maketitle

\begin{abstract}
This paper explores how deliberate mutations of reward function in reinforcement learning can produce diversified skill variations in robotic manipulation tasks, examined with a liquid pouring use case. To this end, we developed a new reward function mutation framework that is based on applying Gaussian noise to the weights of the different terms in the reward function. Inspired by the cost-benefit tradeoff model from human motor control, we designed the reward function with the following key terms: accuracy, time, and effort. The study was performed in a simulation environment created in NVIDIA Isaac Sim, and the setup included Franka Emika Panda robotic arm holding a glass with a liquid that needed to be poured into a container. The reinforcement learning algorithm was based on Proximal Policy Optimization. We systematically explored how different configurations of mutated weights in the rewards function would affect the learned policy. The resulting policies exhibit a wide range of behaviours: from variations in execution of the originally intended pouring task to novel skills useful for unexpected tasks, such as container rim cleaning, liquid mixing, and watering. This approach offers promising directions for robotic systems to perform diversified learning of specific tasks, while also potentially deriving meaningful skills for future tasks.
\end{abstract}


\section{Introduction}
For robots to successfully operate in unstructured and unpredictable real-world environments, they need the ability to constantly adapt and learn many tasks. One way to do this is to learn from human demonstration~\cite{billard2016learning}. However, human involvement can be costly, and humans are not always available to correct or teach robots new skills. Indeed, an alternative is reinforcement learning (RL) that allows the robots to autonomously acquire new skills through trial‑and‑error interaction with their environment~\cite{chatzilygeroudis2019survey}. The robot is given an objective function (typically from a human), which then guides its autonomous exploration to obtain a policy of how to perform a given task. At each timestep, the robot observes the current state, executes an action according to its policy, and receives a scalar reward. Over many episodes, it refines its policy to maximise the expected sum of discounted rewards, thereby acquiring skills optimised for long‑term success.

\begin{figure}[t!]
      \centering
      \includegraphics[width=.98\linewidth]{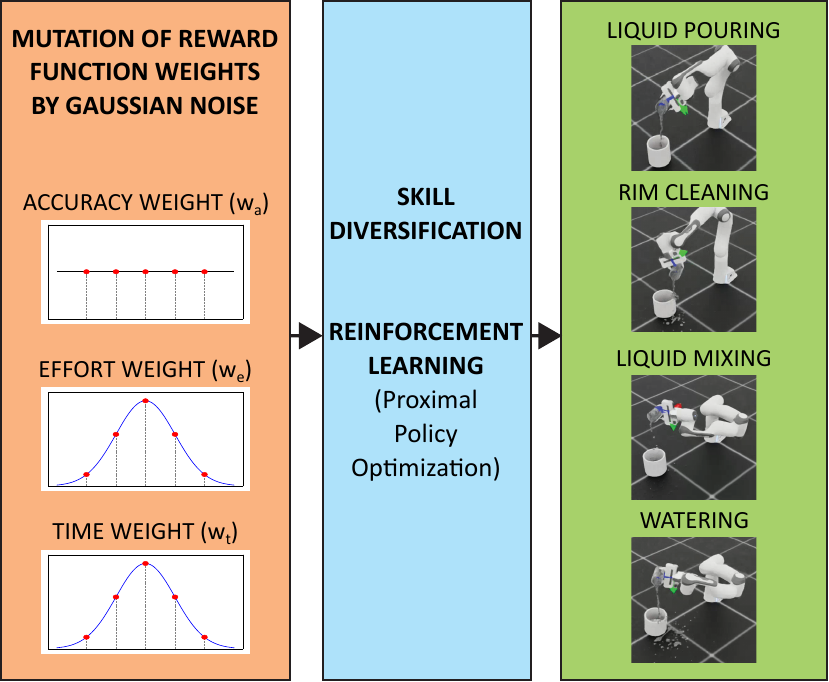}
      \caption{The proposed skill diversification concept with reward function mutation studied on a liquid pouring task. The mutations are induced into the reward function via Gaussian noise distributions (left) on the reward terms (accuracy, time, and effort), which are used to train Proximal Policy Optimization (PPO) policies. The resulting policies exhibit diverse skill variations (right), including fast/slow pouring, rim cleaning, mixing, and watering behaviours.}
      \label{fig:concept}
\end{figure}

RL has been successfully applied to robots to solve a diverse range of tasks, ranging from pick-and-place actions~\cite{franceschetti2021robotic,shahid2022continuous}, object lifting~\cite{roveda2020model}, to assembly~\cite{jin2023container}, as well as play ball-in-the-cup game~\cite{celemin2019reinforcement}, table tennis~\cite{mulling2013learning}, and air hockey \cite{liu2021efficient}. Within this context, the liquid pouring task~\cite{tamosiunaite2011learning,kroemer2012kernel,yamaguchi2016neural,babaians2022pournet} stands out as a particularly compelling benchmark for investigating the role of reward function design in shaping learned behaviours. Unlike binary success criteria seen in stacking or placement tasks, pouring involves balancing multiple continuous objectives, such as avoiding spillage, reducing effort, and maximising efficiency, making it highly sensitive to how learning is incentivised.

RL has achieved impressive results in specific robotic tasks with well-crafted and tailored reward functions. This typically results in good skills specialised for the given task, but lacks generalisation capabilities when new tasks arise. Learning new tasks is typically relatively long and sample-inefficient, especially in complex tasks without prior knowledge and where rewards are sparse or delayed. Rather than relying solely on environmental feedback, an agent can benefit from understanding the reward logic itself, such as temporal dependencies, conditional sequences, or subgoals—thereby improving learning efficiency and policy quality~\cite{kober2011reinforcement,eschmann2021reward,icarte2022reward}. By leveraging structured reward representations, agents can more effectively sequence and reuse behaviours, enabling them to adjust previously learned skills to new or modified tasks. This structured approach facilitates faster adaptation, as agents can generalise from prior experience rather than starting from scratch each time~\cite{pertsch2021accelerating}.

Another approach to reduce learning time and improve generalisability is to utilise direct prior task knowledge from models or human demonstrations~\cite{johannink2019residual,chatzilygeroudis2019survey}. In that way, an agent already has a rough policy, which then only needs to be refined and optimised for the given specifics of the robot and the environment. However, resulting policies that are not considered optimal for the given specific task are often discarded. We argue that such ``failed'' or ``suboptimal'' policies should not be discarded, since such skill variations resulting from various mutations might be useful starting points for learning new tasks.

Mutation of policy in robot learning when subject to physical interaction with humans and unpredictable environments has been observed in~\cite{peternel2017robots}. A follow-up study~\cite{maessen2023robotic} investigated how the policy mutations occur during the learning and what kind of variations of skill emerge in a sawing task. The study concluded that certain policy variations may not be optimal or suitable for the original task, but can be useful for optimising some other parameters/tasks. This highlighted the potential for diversification of skills and the importance of not discarding the policies that appear to be suboptimal for the current task, as they might be a good starting point for unforeseen new tasks. However, the work so far relied on mutations from a passive environment, while intentional mutations with a systematic mechanism are still missing.

To address this challenge, we introduce a reward mutation framework that treats the reward function as a tunable mechanism for active skill diversification (Fig. \ref{fig:concept}). While online reward-shaping has been investigated for improvement of sample efficiency and optimisation of a specific task in~\cite{onori2024adaptive}, differently, the proposed approach mutates the reward function to discover new skills. As a starting point, we study a liquid pouring task performed by a Franka Emika Panda robotic manipulator in the NVIDIA Isaac Sim simulation environment, where the reward is composed of three weighted terms: pouring accuracy, time spent, and effort spent. The goal of the study is to systematically explore how the mutation of the weights of these reward terms induces the emergence of diverse policies. We perform training of agents with Proximal Policy Optimization (PPO) \cite{schulman2017proximal} under 25 distinct reward configurations. Some of the emerging policies are useful for executing the original task in different ways (fast and slow), some are identified to be useful for unforeseen other tasks (container rim cleaning, liquid mixing, watering), and some are not useful for any identifiable tasks.

\section{Methods}\label{methods}

\subsection{Algorithm}
\begin{figure}[t!]
    \centering
    \includegraphics[width=0.98\linewidth]{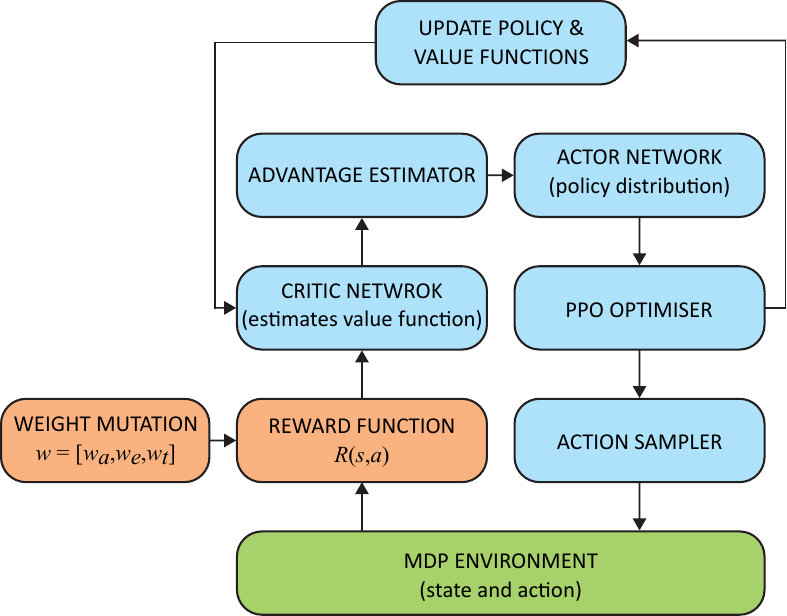}
    \caption{PPO-MDP interaction framework. The PPO agent (blue blocks) interacts with the environment modelled by the Markov Decision Process (MDP) (green), which includes state transitions, actions, and observations. The additional custom reward mechanism (orange) integrates the study’s key terms (accuracy, effort, and time), whose weights are systematically varied to induce mutations of policy.}
    \label{fig:method}
\end{figure}

To enable robotic skill diversification, we developed an RL framework with an intentional reward mutation mechanism. The overview of the framework is shown in Fig.~\ref{fig:method}. For policy learning, we choose the PPO algorithm due to a good balance of stability, performance, and implementation practicality qualities. For details of the PPO algorithm itself, please see \cite{schulman2017proximal}. Here, we will examine the additional intentional reward mutation mechanism.

The reward function is a fundamental element in RL, dictating the incentives that shape the learned policy. Different formulations of the reward function significantly impact the skill adaptation process in robotic learning. In robotic tasks such as pouring, designing a robust reward function is critical because it must balance competing objectives.

In PPO, the reward function directly influences both policy updates and value function approximation. The agent's advantage function is given as
\begin{equation}
A(t) = Q\big(s(t), a(t)\big) - V(s(t)),
\end{equation}
where $Q$ is the discounted sum of rewards and $V$ is the baseline estimate of expected return, while $s(t)$ and $a(t)$ represent state and action, respectively, at given time $t$. The value function is defined as
\begin{equation}
V\big(s(t)\big) = \mathbb{E} \big[ R(t) + \gamma V\big(s(t+1)\big) \big],
\end{equation}
where $\mathbb{E}$ denotes expectation, $R(t)$ is the reward, and $\gamma \in [0,1]$ is the discount factor, which balances immediate and future rewards. The value function is trained to predict cumulative rewards, making reward design pivotal for stable learning. Poorly structured rewards may lead to suboptimal behaviours (e.g., excessive movements or task failure), while well-shaped rewards accelerate policy improvement.

We based the reward function on a cost-benefit tradeoff model from human motor control~\cite{rigoux2012model,berret2016don,peternel2017unifying,shadmehr2019movement}, which accounts for both the cost of the movement in terms of energy expended and the perceived benefit of that movement in terms of time spent on it. The reward was defined as
\begin{equation} 
    R(s,a) = \underbrace{e^{-\frac{t}{w_t}}}_{\text{Time}} \cdot \underbrace{w_a A}_{\text{Accuracy}} - \underbrace{w_e E}_{\text{Effort}},\label{eq:reward}
\end{equation}
where the reward increases with accuracy $A$. In a pouring task, accuracy is measured by the amount of liquid mass transferred into the container and is weighted by $w_a$. Time $t$ spent on the task so far is weighted by $w_t$ and exponentially discounts the reward to ensure that the longer movements are perceived as less beneficial. The underlying reasoning for this is that the agent does not waste time, as executing the task quickly allows for more executions. Finally, the effort term penalises effort $E$ of the action via measured joint torques to prevent excessive energy usage and is scaled by $w_e$. The side product of this is also to achieve smooth trajectories~\cite{flash1985coordination,uno1989formation}.

The proposed concept aims to diversify skills by actively mutating the reward function. To do so, we introduced controlled variations into the reward configuration by applying Gaussian noise to selected reward weights. Starting from a baseline weights of reward function that yields a well-performing pouring strategy, we varied the weights for time ($w_t$) and effort ($w_e$) using a Gaussian noise distribution centred on their respective baseline values. The accuracy-related weight ($w_a$) remained fixed to ensure that the agent had some general objective.

Formally, each mutated reward weight $w_i'$ was defined as
\begin{equation}
    w_i' = w_i + \epsilon_i,\\
    \quad \epsilon_i \sim \mathcal{N}(0, \sigma^2),\label{eq:weight}
\end{equation}
where $w_i$ is the baseline weight value, $\epsilon$ is the Gaussian noise, and $\sigma$ is the standard deviation of the noise that controls the strength of the mutation.

\begin{figure}[t!]
    \centering
    \includegraphics[width=0.98\linewidth]{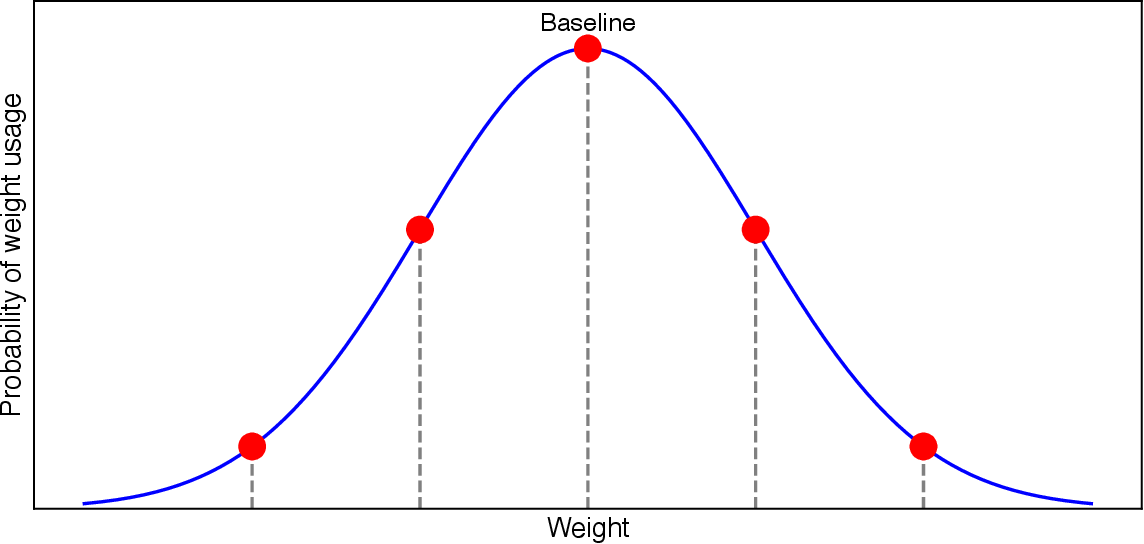}
    \caption{Reward weight sampling on a Gaussian distribution for systematic analysis. The graph depicts the probability density (y-axis) of reward weight configurations (x-axis) sampled from a Gaussian distribution centred on the baseline weights. Five weight configurations for a given term are experimentally tested to probe the reward-function sensitivity.}
    \label{fig:weight_dist}
\end{figure}

To explore how induced mutations in the reward function influence learned behaviour, we examined different values of weights on the Gaussian noise curve. Rather than relying on random sampling from Gaussian noise, we selected four representative values on the Gaussian curve (in addition to the baseline) for each varied weight to systematically examine the effects of such mutations (Fig.~\ref{fig:weight_dist}). To this end, we chose two values below and two above the baseline along the Gaussian distribution, where the baseline was set in the middle of the curve. This structured analysis approach ensured broad yet controlled coverage of the reward mutation space, enabling interpretable and consistent comparisons across policies. If the purpose is not analysis but actual exploration, random Gaussian noise could be used instead.

\subsection{Task Setup}
The liquid pouring task was created in Isaac Lab, which is a reinforcement learning environment built on top of NVIDIA Isaac Sim. The setup included a Franka Emika Panda robotic manipulator that holds a glass with liquid at its end-effector. The robot actions were defined as commanded joint torques. Near the robot was a container where the liquid had to be poured. We placed the container on a digital scale to measure the amount of liquid inside it with force and used the information to determine the accuracy reward. The simulated environment models realistic physical dynamics, including gravity, contact forces, and fluid-like behaviour during the pouring process. The setup can be seen in the results in Figs.~\ref{fig:fast}--\ref{fig:watering}. The task required the robot to pour a controlled amount of liquid into the container by coordinating its end-effector position and orientation. The success of the original task was defined by avoiding spillage, minimising the used energy, and performing the task as quickly as possible. Mutations of the weight in the reward function allowed exploration beyond the original task.

\subsection{Training Procedure}
The core of the learning process is a reward function composed of three weighted terms: force accuracy on the scale, time minimisation, and effort minimisation. A total of 25 distinct weight configurations (a combination of 5 time weights and 5 effort weights) were tested to explore how the mutation of weights across these components influences the resulting policy behaviour. Each reward configuration used in the experiments was selected from a Gaussian distribution centred around a baseline pouring strategy, as illustrated in Fig.~\ref{fig:weight_dist}. For every weight configuration out of 25, we trained three policies to account for stochasticity in learning. Each trained policy was then evaluated across 10 simulations. This resulted in 75 trained policies and 750 total simulations.

\subsection{Data Evaluation}\label{sec:eval}
For the evaluation, we recorded the position and orientation of the end-effector, the z-axis force on the container's scale (used to infer volume poured), the time taken to reach the pouring target, and the actuator effort (joint torque) at each timestep. The end-effector trajectories were normalised relative to the robot’s initial pose. Each policy was manually classified based on observed behaviour according to the following rubric:
\begin{itemize}
    \item Original task: Policy immediately useful for the original pouring task, labelled by the characteristic of execution (e.g., task execution speed).
    \item Potential for other tasks: Policy that failed to complete the original task but demonstrated meaningful behaviour patterns for other tasks.
    \item No policy: Policy could not be linked to the use for any immediately identifiable tasks.
\end{itemize}
Trials in which the pouring process was still ongoing when the evaluation period ended are excluded from the analysis, as goal completion could not be conclusively determined, and the learning could continue indefinitely. This combination of quantitative metrics and qualitative labelling enabled examination of how reward configurations drive behavioural diversity and skill mutation in robotic RL.

\begin{figure}[t!]
    \centering
    \includegraphics[width=1\linewidth]{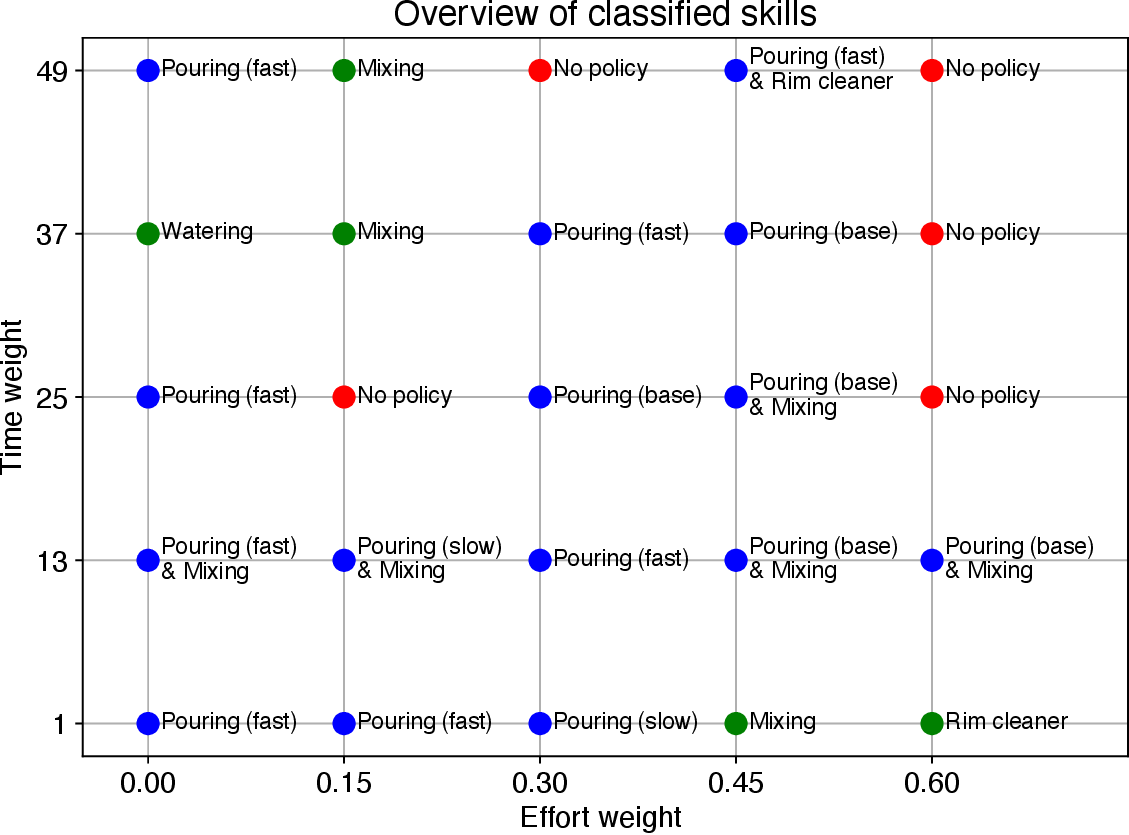}
    \caption{Resulting skill classification by the varied reward weight pairs. Effort weight versus time weight combinations (x/y-axes) are mapped to their resulting skill labels: pouring (slow/base/fast), no policy, mixing, and watering. Points are colour-coded by behaviour class: blue (original goal achieved), green (novel skill but original goal missed), and red (no viable policy). }
    \label{fig:weightskill}
\end{figure}

\section{Results}\label{sec:results}
The overview of the results of the experiments is shown in Fig.~\ref{fig:weightskill}, where we labelled different skill variations. The systematic testing of weight mutations revealed a diverse range of learned pouring behaviours. We classified them according to the approach described in Sec.~\ref{sec:eval}. In the ``original task'' class, we identified three distinct subclasses based on task execution speed: \textit{slow}, \textit{base}, and \textit{fast} pours. In the ``potential for other tasks'' class, we observed several policies that exhibited novel or mutated skills not explicitly trained for. One such emerging skill variation, labelled as \textit{rim cleaning}, was identified to be useful for cleaning the edge of the container as the robot drops the liquid around the container's rim. Another one was labelled as \textit{mixing}, where the liquid is swirled within the container. Finally, in an emerging skill variation, labelled as \textit{watering}, the robot distributes the liquid in a spreading motion similar to watering plants or rinsing a surface.

Out of the 25 trained configurations, 16 policies successfully completed the original pouring task, with some exhibiting slight mutations from the baseline pouring behaviour. Five policies could not be linked to the use for any immediately identifiable tasks. Four produced distinctly new skills for identifiable new tasks, highlighting the potential of active mutations of reward function. In the following sections, we describe and evaluate some of the examples of different policy types that emerged, with the focus on the novel skill variations in response to reward function mutation.

\subsection{Original task: fast/slow pouring}
\begin{figure}[t!]
	\centering
	\subfloat{\includegraphics[width=1.0\columnwidth]{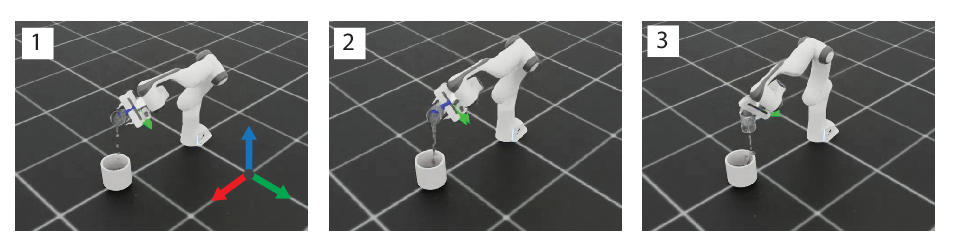}}\\ \vspace{-4mm}
	\subfloat{\includegraphics[width=0.9\columnwidth]{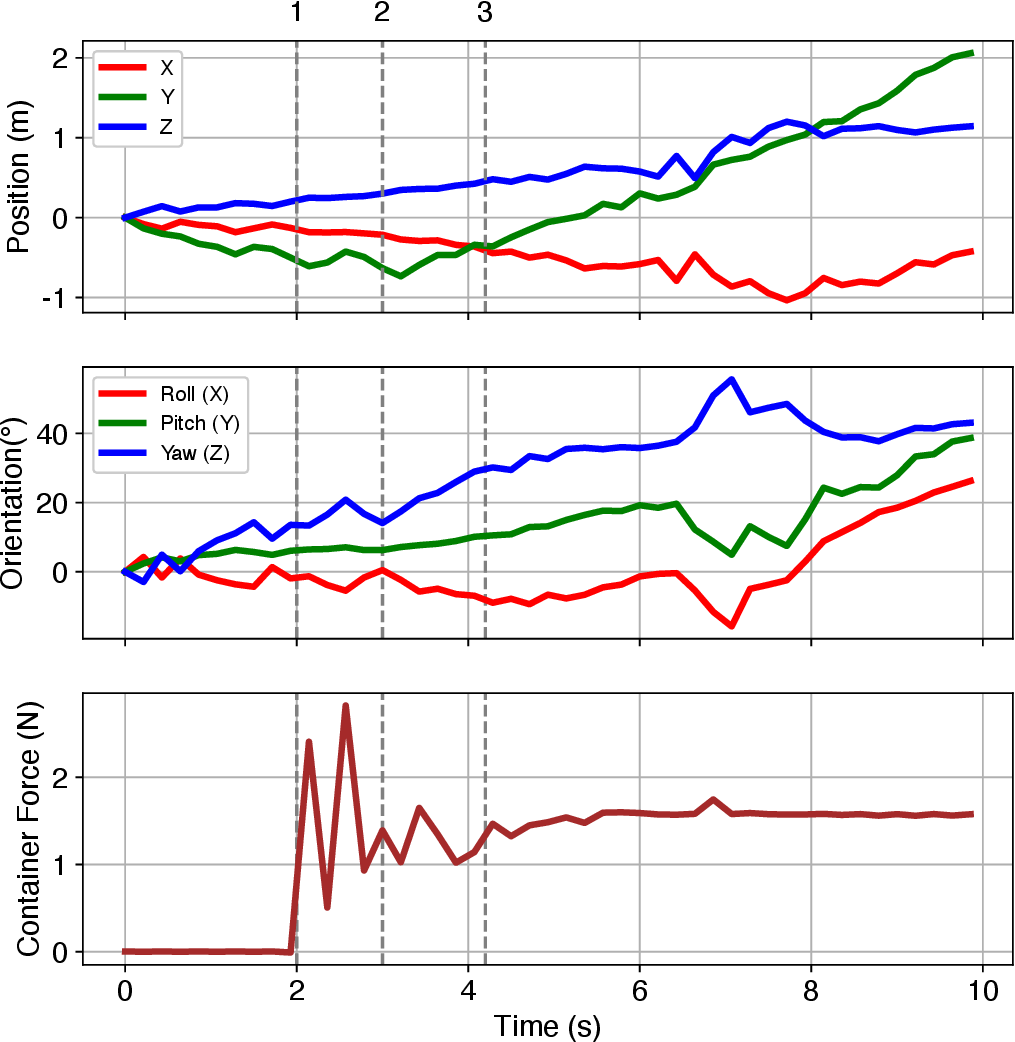}}
    \caption{Fast-pouring skill variation. Storyboard sequence shows the initial, middle, and final stages of the fast pouring. The first graph shows end-effector position (x, y, z), while the second graph shows end-effector orientation (roll, pitch, yaw). The last graph shows the force measured by the scale proportional to the net liquid transfer.}
    \label{fig:fast}
\end{figure}

The fast-pouring policy is shown in Fig.~\ref{fig:fast} and is characterised by a rapid and efficient transfer of liquid into the container. In the first two graphs and in the storyboard above them, we can see that the robot started the trial with a swift tilting movement that accelerates the flow of liquid early in the trial. An interesting aspect can be seen in the y-axis, where the robot first shifted right (negative) and then, after about 3 seconds, moved back to the left (positive), suggesting a lateral correction during pouring. The liquid transfer can be observed from the force data as measured by the scale (third graph), where pouring began around the 2-second mark and reached the desired weight on the scale at approximately 5.5 seconds. Notably, the graph displays a brief force peak at the onset of pouring, caused by the inertia of the initial fluid release, which is followed by several oscillations as the liquid pours in.

In general, the slow-pouring policy is similar to the fast-pouring policy but with a slower execution time. For the sake of space, we show only graphs for the fast-pouring policy and examine completion time for comparison. The slow-pouring policy managed to fill the container in about 7.5 seconds, compared to 5.5 seconds.

\subsection{Rim cleaner}
\begin{figure}[t!]
	\centering
	\subfloat{\includegraphics[width=1.0\columnwidth]{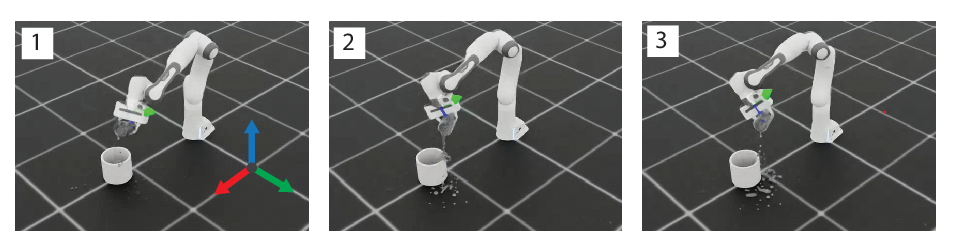}}\\ \vspace{-4mm}
	\subfloat{\includegraphics[width=0.9\columnwidth]{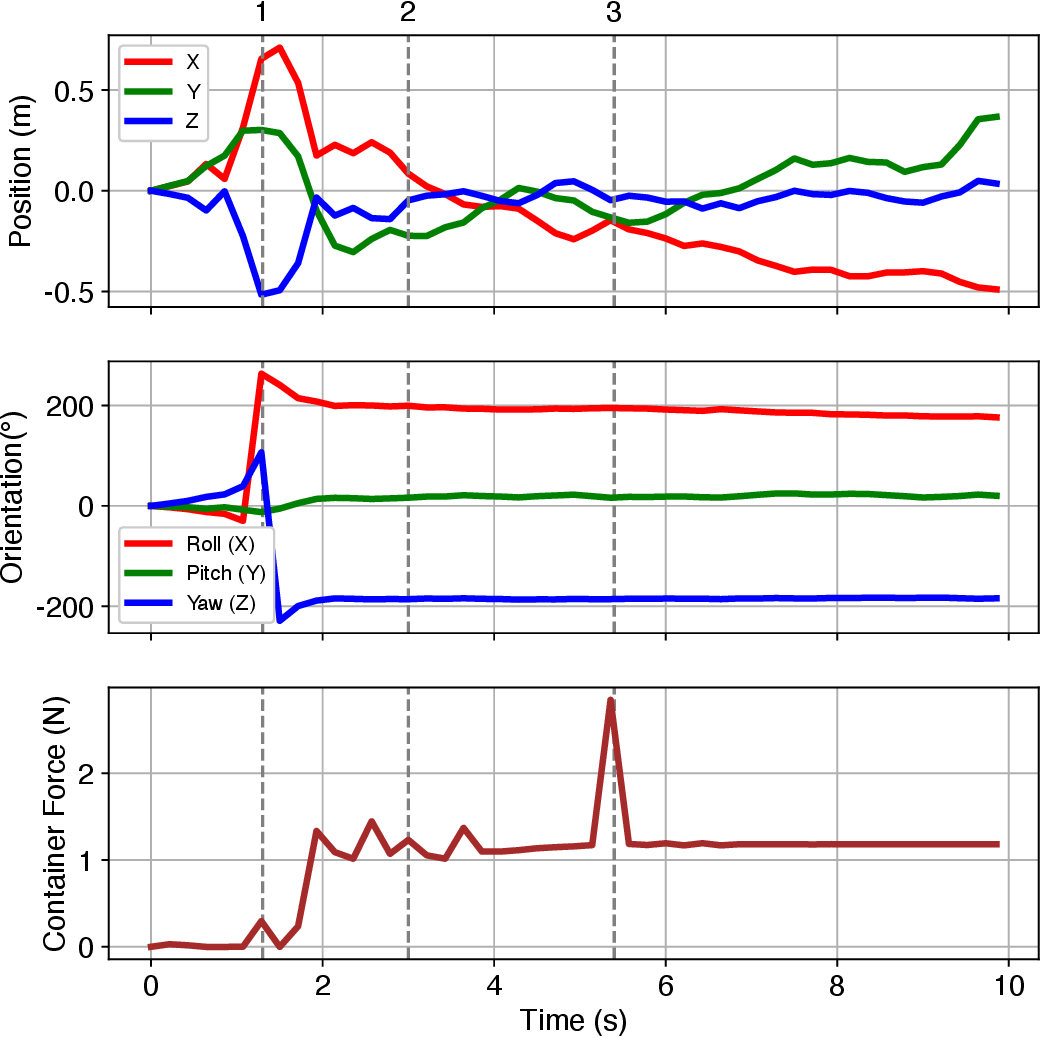}}
    \caption{Rim-cleaner skill variation. Storyboard sequence shows the initial, middle, and final stages of the rim cleaning behaviour. The first graph shows end-effector position (x, y, z), while the second graph shows end-effector orientation (roll, pitch, yaw). The last graph shows the force measured by the scale proportional to the net liquid transfer.}
    \label{fig:rim}
\end{figure}

The rim-cleaner policy is shown in Fig.~\ref{fig:rim} and exhibits a distinct deviation from the originally intended pouring task. The liquid was poured on the edge of the container instead of inside it, which could be useful to clean the rim of the container. The pouring of liquid on the rim can be clearly seen on the storyboard above the graphs. According to the position data (first graph), the x-axis shows a gradual movement back toward the robot base, while the y-axis initially shifted to the left before correcting toward the right during pouring (from 1 to 6 seconds). The z-axis remained relatively constant, indicating that the pouring occurs at a fixed height.

From the force data measured by the scale, we can see that the container was still partially filled, since some of the liquid went inside and some outside. However, it is important to note that this skill did not pour into the container consistently. In some trials, the robot poured partially or entirely on the outer edge of the container. Despite this, the rim-cleaner policy demonstrates a potentially valuable and adaptable skill that could be repurposed for tasks involving targeted surface interaction or cleaning motions, even if it falls short in terms of the original task objective.

\subsection{Liquid Mixing}
\begin{figure}[t!]
	\centering
	\subfloat{\includegraphics[width=1.0\columnwidth]{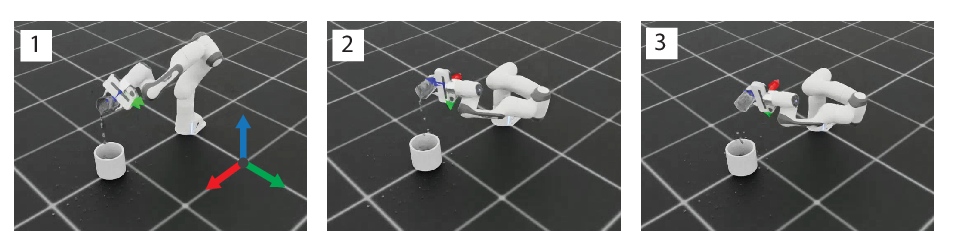}}\\ \vspace{-4mm}
	\subfloat{\includegraphics[width=0.9\columnwidth]{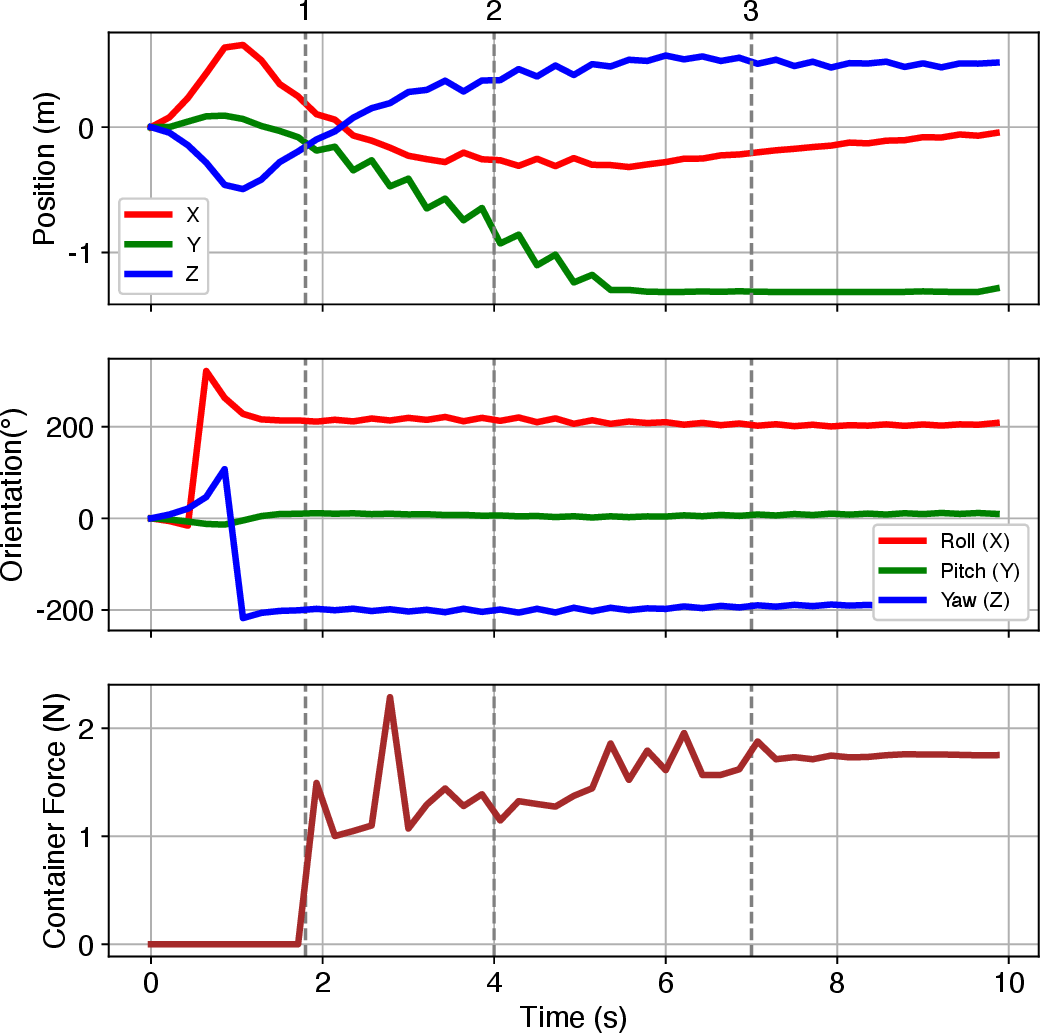}}
    \caption{Liquid-mixing skill variation. Storyboard sequence shows the initial, middle, and final stages of the liquid mixing behaviour. The first graph shows end-effector position (x, y, z), while the second graph shows end-effector orientation (roll, pitch, yaw). The last graph shows the force measured by the scale proportional to the net liquid transfer.}
    \label{fig:mixing}
\end{figure}

The liquid-mixing policy is shown in Fig.~\ref{fig:mixing}, which demonstrates a unique pouring strategy characterised by rhythmic shaking movements during liquid transfer. The shaking aspect is clearly visible in the position graph (first graph), in particular in the y-axis. The y-axis position decreased steadily, indicating a rightward shift, but with noticeable oscillations during pouring, suggesting a shaking motion layered on top of a gradual lateral movement. On the z-axis, the end-effector initially lowered and then, after the first second, gradually lifted during the pour, again with small fluctuations visible in the trajectory, reinforcing the interpretation of a deliberate vibratory motion. These small but consistent oscillations resulted in a dynamic motion that can effectively stir or mix the liquid as it is poured.

The shaking behaviour produced a stepped or pulsed increase of the liquid, as seen in the force data from the weight (third graph). This indicates that the liquid entered the container in bursts rather than a continuous stream. Despite this unsteady pattern, the robot successfully reached the pouring goal in this instance, with only a small amount of spillage. However, across multiple trials, this skill was more prone to overshoot or spill due to the inherent instability of the shaking motion. Still, this behaviour demonstrates an interesting mutation of the originally intended policy and is potentially valuable for applications where simultaneous mixing and pouring are desired.

\begin{figure}[t!]
	\centering
	\subfloat{\includegraphics[width=1.0\columnwidth]{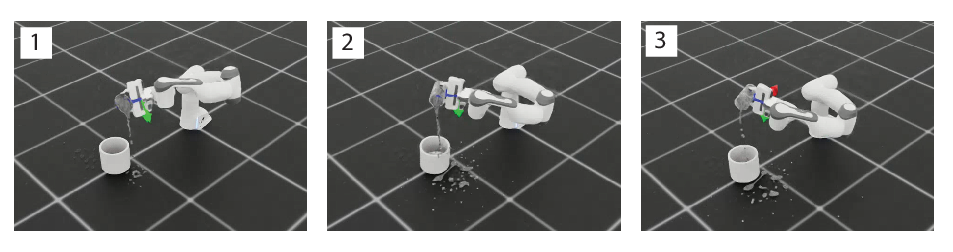}}\\ \vspace{-4mm}
	\subfloat{\includegraphics[width=0.9\columnwidth]{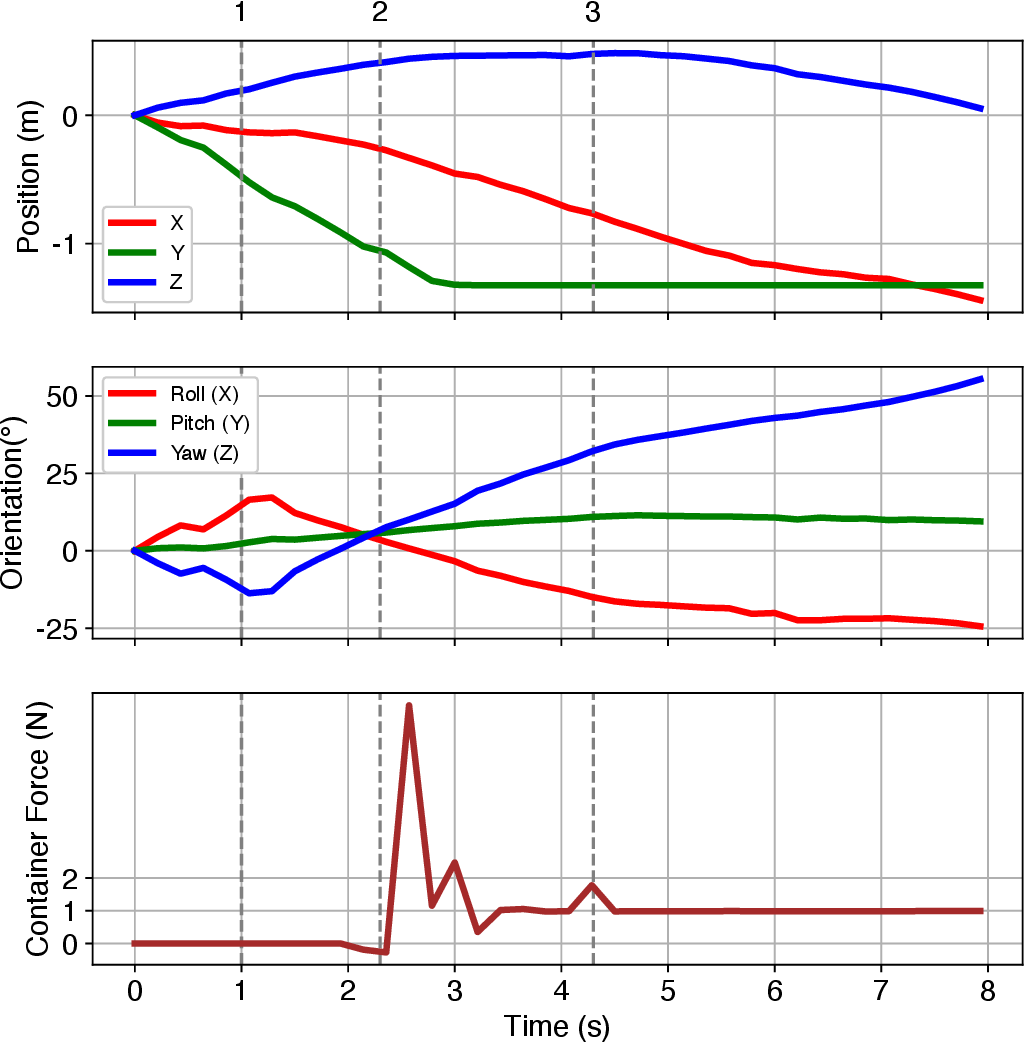}}
	\caption{Watering skill variation. Storyboard sequence shows the initial, middle, and final stages of the watering behaviour. The first graph shows end-effector position (x, y, z), while the second graph shows end-effector orientation (roll, pitch, yaw). The last graph shows the force measured by the scale proportional to the net liquid transfer.}
	\label{fig:watering}
\end{figure}

\subsection{Watering behavior}
The watering policy is shown in Fig.~\ref{fig:watering} and represents the most distinct deviation from the originally intended pouring task that was observed in this study. Rather than focusing on efficient and accurate transfer of liquid into the container, this skill dispersed the liquid broadly across an area, resembling the motion one might use to water a garden bed or rinse a surface. While observing the storyboard, we can see that the robot performed a sweeping motion that resulted in a splash. The orientation data (second graph) highlights the dynamic nature of this policy. The rotation around the x-axis first increased up to 20 degrees and then gradually decreased to around -20 degrees, while the z-axis followed the opposite pattern, which resulted in a coordinated twisting motion during the pouring. The y-axis orientation remained near its initial position, indicating that the rotation was roughly confined to a single plane.

This twisting trajectory caused the robot to begin pouring outside of the container’s bounds and continue moving away from it during execution. In the force data measured by the scale (third graph), we can see a sharp spike at the 2.5-second mark under the impact of liquid flow. The force reading then stabilised around 1 N, which was insufficient to meet the original task objective. Due to a large area onto which the liquid was spread, some liquid went into the container while some went outside. This emerging policy may be applicable in tasks requiring broad liquid distribution, such as watering plants, rinsing surfaces, or distributing cleaning agents over a wide area.

\section{Discussion and Conclusions}
The goal of the study was to introduce a reward mutation framework for active skill diversification and to systematically explore how the mutation of the weights of these reward terms induces the emergence of diverse policies. Our idea was inspired by the theory of evolution, where random mutations produce variations in species. These variations may not be optimal for the current conditions; however, they can be a possible good fit for future changed conditions. In the same sense, actively mutating robot behaviour can produce variations of skills that might not be suitable for the existing tasks, but may be a good fit for another unexpected task or at least a starting point for learning it.

The proposed approach is different from classical RL exploration. In classical exploration, the noise is induced into actions, while the proposed approach induces noise into reward weights. In this sense, the mutation of specific weights provides more targeted exploration than randomising actions themselves.

The study revealed that actively mutating the reward function mostly produced variations of skill suitable for the original task (16 out of 25 tested configurations), four configurations gave variations useful for unexpected tasks, and five were not identified as useful. One could argue that this gives a relatively good proportion in terms of exploration, since most of the time invested is still spent on the intended task, while less time is spent on finding variations for new tasks. Indeed, when employing this strategy, one must consider the trade-off between the discovery of new skills and time spent on learning/exploring.

Besides finding skill variations of the original pouring task, we were able to discover three new variations not fit for the original task but suitable for unexpected new tasks. One skill exhibited a behaviour where the robot poured the liquid around the edge of the container, which can be beneficial for cleaning the rim. Another skill manifested in a behaviour where the robot was shaking the glass while pouring, which could be useful when the liquid has to be mixed. We also identified a skill that exhibited a behaviour where the robot was splashing the liquid over a distributed area, which can be beneficial for watering plants or rinsing a surface.

One of the key challenges is labelling the emerging variations, where the labelling was performed manually in this study. For a more effective and wider application of the proposed approach, a method for automatic labelling needs to be developed. In that direction, visual language models (VLMs) could play a pivotal role in detecting the context from human language and data that was trained on videos/pictures of daily tasks  \cite{li2023vision,lai2024vision,jekel2025visio}. The combination of foundation models and RL is a promising research topic in robotics, especially considering RL reward design \cite{moroncelli2024integrating}.

Another future work direction is to explore the proposed approach also in other tasks. While liquid pouring is a very challenging task and thus a good use case for this initial study, the ratio between the three classes of skill variations (i.e., original task, new task, and not useful) could differ between different tasks. The experiments and learning could also be potentially performed on a real-world setup. Nevertheless, the sim-to-real robotic skill learning approach has proven a very successful alternative to time-consuming, expensive, and potentially dangerous learning on a real-world setup~\cite{zhao2020sim}.

In future work, this framework could be extended with multi-objective reinforcement learning techniques to better balance trade-offs across competing goals. Alternatively, human-in-the-loop feedback mechanisms could help guide behaviour toward more socially or operationally preferable solutions. Applying this reward mutation concept to new tasks such as stirring, scooping, or tool use would help generalise the methodology, while transitioning to physical robots would validate the approach under real-world dynamics and constraints.

\bibliographystyle{IEEEtran}
\small
\setstretch{0.0}
\bibliography{preprint}

\end{document}